\documentclass[11pt]{article}

\usepackage[margin=1in]{geometry}
\usepackage{amsmath}
\usepackage{array}
\usepackage{booktabs}
\usepackage{caption}
\usepackage{enumitem}
\usepackage{float}
\usepackage{graphicx}
\usepackage{listings}
\usepackage{microtype}
\usepackage{needspace}
\usepackage[numbers,sort&compress]{natbib}
\usepackage{placeins}
\usepackage{xcolor}
\usepackage{hyperref}
\usepackage{tabularx}

\definecolor{draftblue}{HTML}{1F4E79}
\definecolor{draftgray}{HTML}{F3F5F7}
\definecolor{promptbg}{HTML}{F7F8FA}
\definecolor{promptborder}{HTML}{CBD5E1}
\hypersetup{colorlinks=true,linkcolor=draftblue,citecolor=draftblue,urlcolor=draftblue}

\newcolumntype{Y}{>{\raggedright\arraybackslash}X}
\lstdefinestyle{prompt}{
  basicstyle=\ttfamily\footnotesize,
  backgroundcolor=\color{promptbg},
  frame=single,
  rulecolor=\color{promptborder},
  breaklines=true,
  columns=fullflexible,
  keepspaces=true,
  showstringspaces=false,
  xleftmargin=0.8em,
  xrightmargin=0.3em,
  aboveskip=0.45em,
  belowskip=0.8em
}

\title{\textbf{PolyComp: A Polycube-based Benchmark for Compositional 3D Spatial Reasoning in Multimodal Models}}
\author{Siddharth Patel}
\date{August 9, 2026}

\begin{document}
\maketitle

\begin{abstract}
We introduce \textsc{PolyComp}, a procedurally generated and verified
benchmark that stresses visual recognition and compositional spatial reasoning.
In each problem, a model must identify which of four options shows a pair of
polycube components that can be combined to form a target solid. The benchmark
contains 120 problems across four geometry families, and each problem has three
different presentation formats using either a single image or multiple images. The random guessing baseline is 25\%. Across the three presentations (360
presented problems per model), GPT-5.6 Sol with max effort attains 50.0\%
accuracy (95\% problem-cluster CI 43.3--56.7\%) at a mean cost of \$0.951 per
presented problem, Claude Fable 5 with max effort attains 39.4\%
(33.1--46.1\%) at \$0.701, and Gemini 3.1 Pro Preview with thinking level high
attains 27.5\% (22.8--32.5\%), near the 25\% random guessing baseline, at \$0.350. The observed accuracy spread across geometry families is larger than across presentation formats. We present a problem development and evaluation protocol, cost and token accounting, and release the 120 problems.
\end{abstract}

\section{Introduction}

Spatial intelligence comprises many capabilities spanning perception (e.g., segmentation, entity detection, identification) and reasoning (e.g., reconciling viewpoints and occlusion, geometric transformations). While
multimodal artificial intelligence models have successfully incorporated image
processing to extend the considerable capabilities of large language models
(LLMs) into new domains \citep{yue2024mmmu,lu2024mathvista}, it remains unclear
whether LLM-based models can achieve human-level performance on visual
tasks. Targeted diagnostics continue to find failures in visual and spatial
reasoning \citep{fu2024blink,tong2024eyes,yang2025vsi}. In particular,
human-level performance in spatial reasoning may require the ability to develop
an internal representation of an object and simulate transformations on that
representation.

These abilities have been studied through mental rotation
\citep{shepard1971mental} and qualitative spatial reasoning
\citep{cohn2008qualitative}. They remain difficult for modern vision-language
models, even when the models perform strongly on broad multimodal evaluations
\citep{fu2024blink,tong2024eyes,yang2025vsi}.

Existing benchmarks span synthetic compositional reasoning
\citep{johnson2017clevr}, natural-image relations
\citep{yang2019spatialsense,liu2023vsr}, expert multimodal questions
\citep{yue2024mmmu,lu2024mathvista}, and 3D scene or video understanding
\citep{azuma2022scanqa,ma2023sqa3d,yang2025vsi}. \textsc{PolyComp} targets a
complementary regime: geometric object perception, rigid body transformation,
and composition. The use of polycubes makes the geometry discrete, so that the reasoning task can be solved by identifying a Cartesian frame and counting blocks in three directions. Whether multimodal models build an internal representation
that supports this type of reasoning is unknown.

\Needspace{10\baselineskip}
The contributions of this paper are:

\begin{itemize}[]
\item a procedure to generate and validate a polycube decomposition problem;
\item a comparison of the performance of three frontier multimodal models on 120 of these problems across three presentations, including accuracy and cost efficiency;
\item the 120 problems themselves.
\end{itemize}

\section{Related Work}

\subsection{General multimodal evaluation}

MMMU and MathVista evaluate deliberate reasoning over heterogeneous visual
materials \citep{yue2024mmmu,lu2024mathvista}. MM-Vet emphasizes integrated
capabilities and open-ended grading \citep{yu2024mmvet}; in contrast,
\textsc{PolyComp} uses multiple-choice responses and block geometry. MMStar and MMMU-Pro explicitly address text-only shortcuts, data leakage,
and robustness to presentation changes \citep{chen2024mmstar,yue2024mmmupro}.
Those concerns motivate our decision to present the same problem in three different presentations.

\subsection{Vision-centric and spatial benchmarks}

BLINK reformulates classic computer-vision tasks, including multi-view
reasoning, for multimodal models \citep{fu2024blink}. CV-Bench evaluates 2D
relations and counting as well as 3D depth and distance
\citep{tong2024cambrian}. VSI-Bench tests spatial memory from video and finds
that explicit cognitive maps can help where linguistic chains of thought do
not \citep{yang2025vsi}. InternSpatial-Bench broadens single- and multi-view
instruction formats \citep{deng2025internspatial}; iVISPAR adds interactive
planning \citep{mayer2025ivispar}; SpatialBench proposes a hierarchy from
perception to planning \citep{xu2025spatialbench}. \textsc{PolyComp} is
narrower: it isolates whether models can determine exact
part-to-whole compatibility under 3D rotation for highly constrained polycube geometry.

\subsection{Compositional and 3D reasoning}

CLEVR established synthetic, programmatically controlled visual diagnostics
\citep{johnson2017clevr}; GQA extended compositional questions and functional
programs to natural images \citep{hudson2019gqa}; and NLVR2 tested paired-image
reasoning with compositional language \citep{suhr2019nlvr2}. SpatialSense, VSR,
and Winoground show persistent failures on relations, reference frames, and
minimal compositional contrasts
\citep{yang2019spatialsense,liu2023vsr,thrush2022winoground}. ScanQA, SQA3D,
and 3D-LLM ground questions and language models in explicit 3D scenes
\citep{azuma2022scanqa,ma2023sqa3d,hong2023threedllm}. Our problem images remove scene semantics and external knowledge, concentrating evaluation on
shape, topology, and rigid transformation. Our use of polycubes is inspired by the \texttt{SpatialBench} repository \citep{jaff2026spatialbenchrepo}.

\begin{figure}[H]
  \centering
  \includegraphics[width=\linewidth]{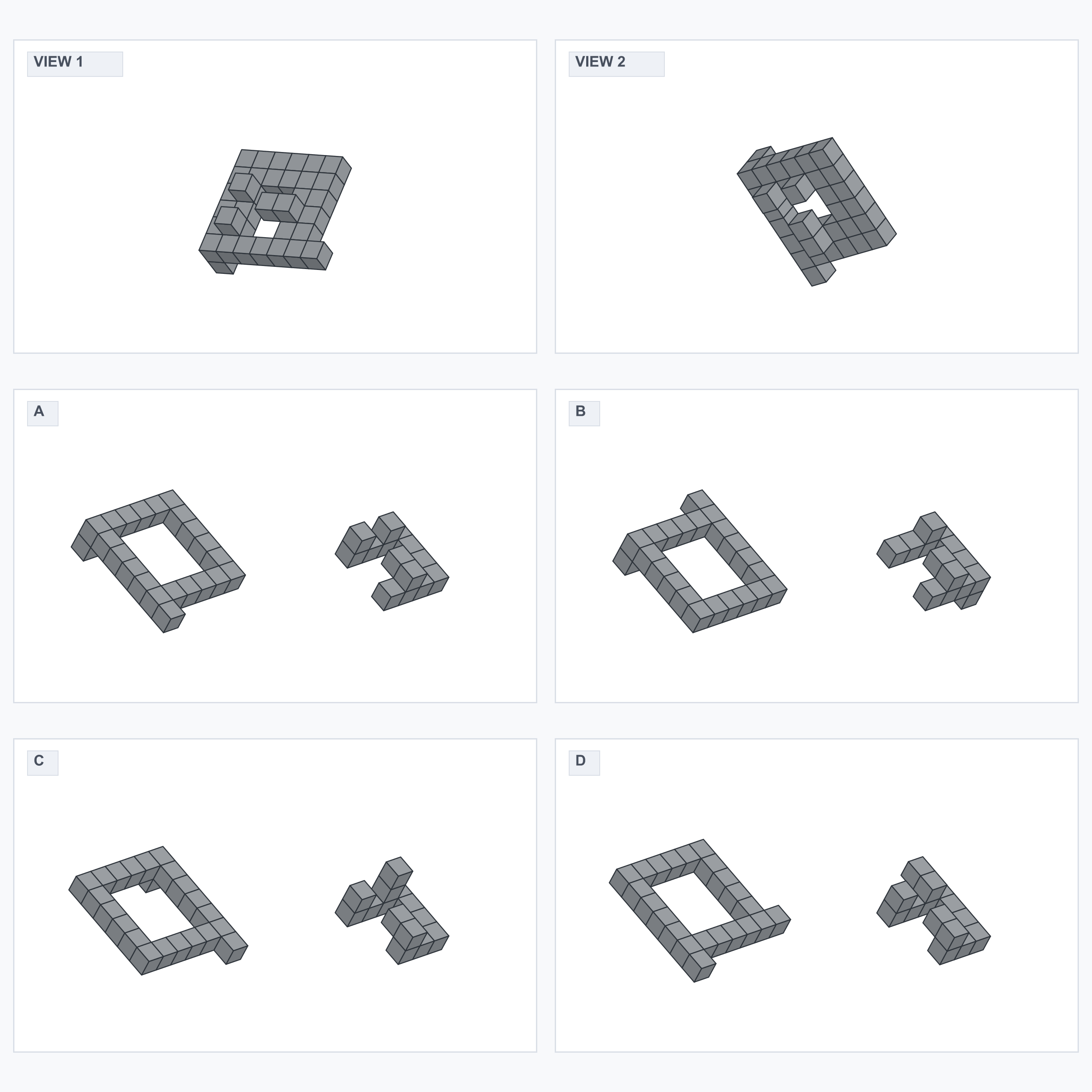}
  \caption{\textbf{Example PolyComp problem.} The model is given the two views of the target solid and then asked to identify which of options A through D shows two components that together can form the target solid. This is one of the simpler problems missed by the three evaluated models, all of which picked D instead of the correct option A.}
  \label{fig:example}
\end{figure}
\FloatBarrier

\section{The PolyComp Benchmark}

\subsection{Problem}

Each problem consists of a single target solid polycube presented in two views, followed by four options labeled A through D, each showing two disconnected polycube components.
Only one option
contains components whose voxel union can equal the target after independent rigid rotations and translations. The model is asked to return one label from
$\{A,B,C,D\}$.

\subsection{Generation and exact verification}

Instances are generated as integer-coordinate voxel sets: the whole target solid and a partition of that solid into two components, which are used as the correct option for the multiple-choice decomposition question. The incorrect options are generated by transforming the correct components by moving a cube or by replacing a component with its mirror image. These transformations preserve cube count while violating the assembly condition. A verification routine enumerates the 24 orientation-preserving cube rotations and feasible integer translations, and confirms that exactly one option can form the target without overlaps or missing cubes. The views for the target solid and the components are selected deterministically from sampled camera rotations. We confirm that every incorrect option can be ruled out from the rendered views.

\begin{figure}[H]
  \centering
  \includegraphics[width=\linewidth]{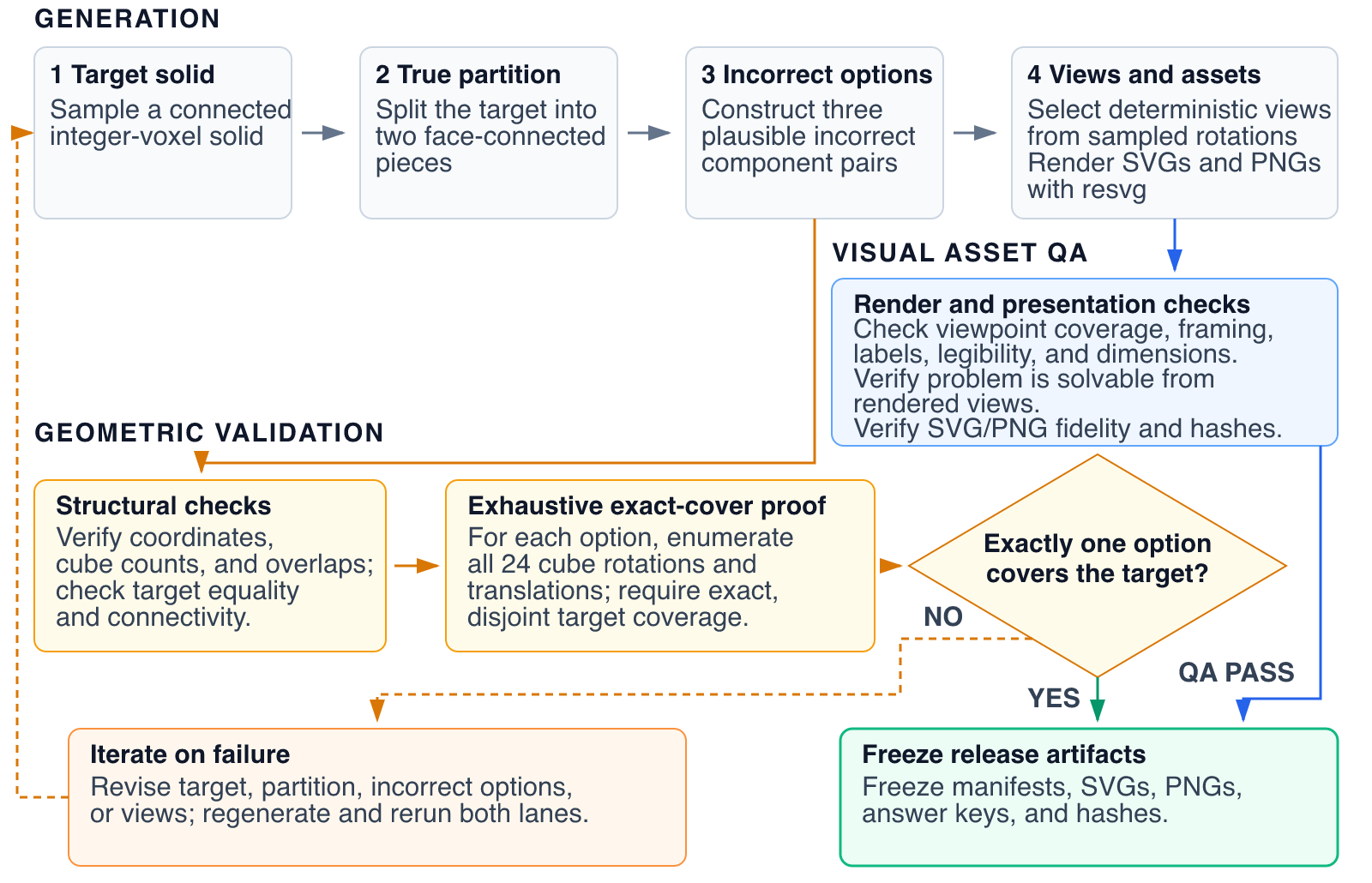}
  \caption{\textbf{Problem generation and validation.} The flow chart shows
  target solid construction, partitioning, incorrect option generation,
  view selection, geometry verification, rasterization, and asset QA.}
  \label{fig:pipeline}
\end{figure}

\subsection{Families and presentations}

The set of 120 problems contains 65 cases based on rectangular loops, 24 cases inspired by toys, 14 cases based on block cleavage, and 17 inspired by joinery. Every problem is presented in three ways:
\newpage
\begin{itemize}
\item a single image with views of the target solid and the options, as in Figure \ref{fig:example}, with the accompanying prompt:
\begin{lstlisting}[style=prompt]
The image above shows two views of the same target solid in the top row. The four options A through D appear below the top row. Each option shows two component solids.
Which option shows two component solids that together can form the target solid?
Return only a JSON object with exactly one key and no additional text:
- "option": one of "A", "B", "C", or "D"
\end{lstlisting}

\item a multi-image presentation in which one image has two views of the target solid and four more images each have views of an option, with non-descriptive model-facing labels \texttt{Image 1} through \texttt{Image 5};
\begin{lstlisting}[style=prompt]
Five images labeled Image 1 through Image 5 appear above. Image 1 shows two views of the same target solid. Image 2 through Image 5 show option A through option D, respectively. Each option image shows two component solids.
Which option shows two component solids that together can form the target solid?
Return only a JSON object with exactly one key and no additional text:
- "option": one of "A", "B", "C", or "D"
\end{lstlisting}

\item a multi-image presentation using the same images as above but with descriptive model-facing labels \texttt{Target} and \texttt{Option A} through \texttt{Option D}.
\begin{lstlisting}[style=prompt]
Five images labeled Target and Option A through Option D appear above. The Target image shows two views of the same target solid. Each option image shows two component solids.
Which option shows two component solids that together can form the target solid?
Return only a JSON object with exactly one key and no additional text:
- "option": one of "A", "B", "C", or "D"
\end{lstlisting}
\end{itemize}

The intent of varying the formulation is to determine whether a given presentation makes perception easier and enables improved spatial reasoning performance.

The single image is generated as vector graphics (SVG) on a 2500$\times$2500 canvas and rasterized to PNG with \texttt{resvg 0.47.0} for submission to the model APIs, without resizing or downsampling. The multi-image target (2440$\times$720) and option images (1200$\times$720) are lossless crops from the same PNG.

\subsection{Selection and scope}

The 120 problems are a subset of 168 submitted in ChatGPT temporary chats using GPT-5.5 at Extra High. The subset comprises 100 problems GPT-5.5 missed and the 20 that it got correct but took the longest time to complete. It is therefore a selected \emph{challenge set}, not a representative sample of spatial tasks or of the problem generation process's unconditional distribution.

\section{Evaluation Protocol}

We evaluate GPT-5.6 Sol with max effort, Claude Fable 5 with max effort, and Gemini 3.1 Pro Preview with high thinking level. Each model receives the 120 problems in each of the three presentations. We record one answer per model and compute pass@1 accuracy. A single Claude pre-output refusal remains in the denominator.

\begin{table}[H]
\centering
\caption{\textbf{Protocol.} Dates identify the evaluation
window. Prices are regular API list rates per million tokens, captured July 16
and rechecked July 25, 2026. Sampling parameters were omitted.}
\label{tab:protocol}
\scriptsize
\setlength{\tabcolsep}{3pt}
\begingroup
\renewcommand{\arraystretch}{1.25}
\begin{tabularx}{\linewidth}{@{}>{\raggedright\arraybackslash}p{0.14\linewidth}YYY@{}}
\toprule
Field & OpenAI & Anthropic & Gemini \\
\midrule
Endpoint and mode & Batch wrapper around \texttt{POST /v1/responses} & \texttt{/v1/messages/batches} & \texttt{:batchGenerateContent} \\
Model; dates & \texttt{gpt-5.6-sol}; July 20--25 & \texttt{claude-fable-5}; July 20 & \texttt{gemini-3.1-pro-preview}; July 20 \\
Reasoning and output cap & \texttt{reasoning.effort=max}; 128,000 tokens & \texttt{output\_config.effort=max}; managed thinking; 128,000 tokens & \texttt{thinkingLevel=high}; 65,536 tokens \\
Image accounting & base64 PNG data URL; \texttt{detail=original} & base64 image source; provider-managed resolution & base64 PNG \texttt{inline\_data}; \texttt{MEDIA\_RESOLUTION\_HIGH}\\
Regular list rates & input \$5; cached \$0.50; cache write \$6.25; output \$30 & input \$10; cache read \$1; 5m/1h write \$12.50/\$20; output \$50 & input \$2; cached \$0.20; output (including thinking) \$12 \\
\bottomrule
\end{tabularx}
\endgroup
\end{table}

Because each problem appears in all three presentations, the 360 outcomes per model are not independent observations. The reported 95\% intervals resample the 120 problem identifiers and retain all presentations for each sampled problem \citep{field2007bootstrapping}. Presentation-specific intervals use the same problem bootstrap, while presentation contrasts use paired resampling of the two outcomes for each problem.

\section{Results}

\subsection{Overall accuracy and efficiency}

\begin{table}[H]
\centering
\caption{\textbf{Primary results.} Confidence intervals are 95\%
problem-cluster percentile-bootstrap intervals over 120 problems, with all
presentations for a sampled problem resampled together. Token summaries are per final response. Costs are
normalized to regular, non-batch API list prices for every provider.}
\label{tab:main}
\small
\begin{tabularx}{\linewidth}{Xrrrrrr}
\toprule
Model & Accuracy & 95\% CI & Mean tok. & Median tok. & Mean cost & Median cost \\
\midrule
GPT-5.6 Sol & 50.0\% & [43.3, 56.7] & 42,682 & 40,549 & \$0.951 & \$0.881 \\
Claude Fable 5 & 39.4\% & [33.1, 46.1] & 19,016 & 17,961 & \$0.701 & \$0.634 \\
Gemini 3.1 Pro Preview & 27.5\% & [22.8, 32.5] & 32,566 & 30,430 & \$0.350 & \$0.329 \\
\bottomrule
\end{tabularx}
\end{table}

\begin{figure}[!t]
  \centering
  \includegraphics[width=0.9\linewidth]{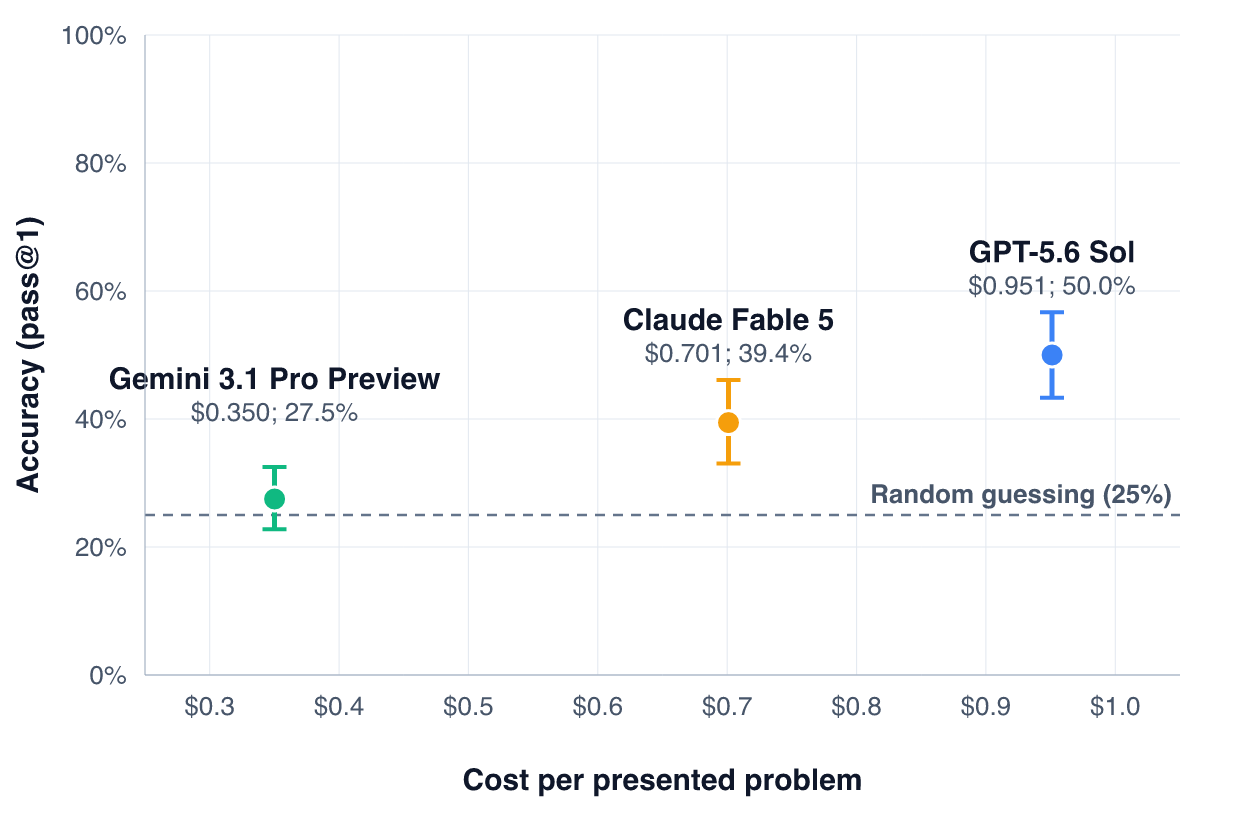}
  \caption{\textbf{Accuracy versus cost.} Each
  point summarizes one model over 360 problem-presentation responses. The
  x-axis is mean list-price cost per presented problem; vertical whiskers are 95\% problem-cluster bootstrap
  intervals. The dashed line is the 25\% uniform random-guessing baseline.}
  \label{fig:cost}
\end{figure}

\begin{figure}[!t]
  \centering
  \includegraphics[width=\linewidth]{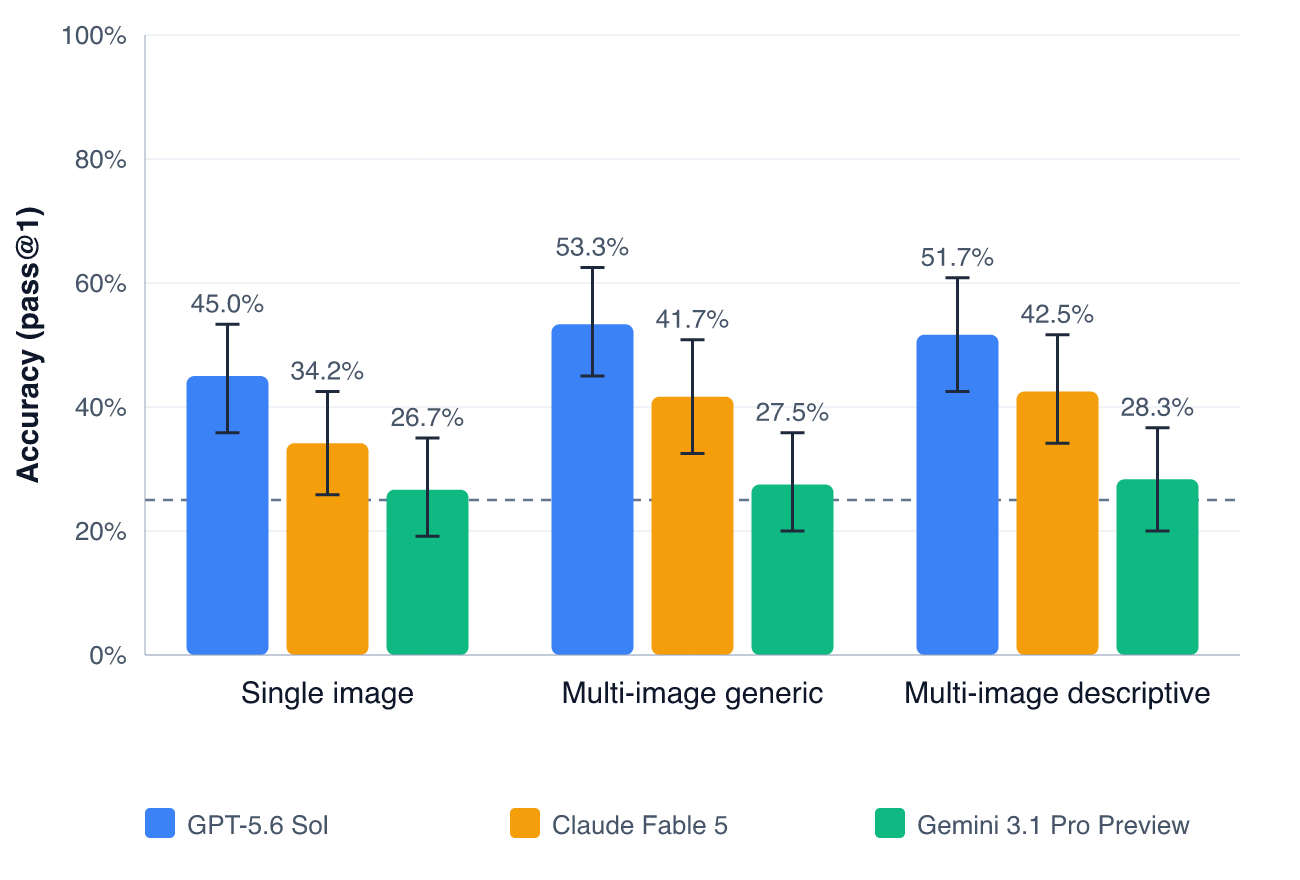}
  \caption{\textbf{Accuracy by model and presentation formulation.} Whiskers
  are pointwise 95\% problem-bootstrap intervals over 120 problems. The dashed
  line marks the 25\% random guessing baseline. There is weak evidence that the multi-image presentations improved performance for GPT-5.6 Sol and Claude Fable 5.}
  \label{fig:accuracy}
\end{figure}

GPT-5.6 Sol leads at 50.0\% accuracy, followed by Claude Fable 5 at 39.4\% and
Gemini 3.1 Pro Preview at 27.5\% (Table~\ref{tab:main}). The uniform random-guessing baseline is 25\%. Using normalized regular API list rates, the complete 360-response evaluations cost an
estimated \$342.41 for GPT-5.6 Sol, \$252.36 for Claude Fable 5, and \$126.06
for Gemini 3.1 Pro Preview. The corresponding mean costs per response are
\$0.951, \$0.701, and \$0.350.

\subsection{Presentation and geometry}

GPT-5.6 Sol's point estimate rises from 45.0\% with a single composite image to 53.3\% with
separate generic images; Claude's rises from 34.2\% to 42.5\% with separate
descriptive images. Gemini remains between 26.7\% and 28.3\%. Paired
problem-bootstrap intervals include zero for every separated-minus-single
contrast. The pattern is consistent with the possibility that
breaking up the single image reduces the perception burden (segmentation,
option identification), but that mechanism was not measured.

The observed spread across geometry families is larger than across presentation formats. GPT-5.6 Sol reaches
81.0\% on block cleavage and 74.5\% on joinery, but 34.4\% on rectangular loop cases. Claude shows the same ordering at 69.0\%, 56.9\%, and 28.7\%. One possible explanation is that the block cleavage and joinery cases start with rectangular target solids, making some incorrect components easier to rule out when they are missing a corner or have an extra block that exceeds the target solid's maximum dimension.

\subsection{Shared successes and failures}

On 24 of the 360 problem-presentation cells, all three models answer correctly, and they all answer incorrectly on 91. At the problem level, eight problems are missed in all presentations by all models. Based on qualitative examination, shared successes tend to have diagnostic global silhouettes or large complementary interfaces, while shared failures more often require tracking small cavities, hidden contact surfaces, or multiple locally plausible alignments. These interpretations remain tentative hypotheses for the time being.

\section{Release}

The public GitHub repository at \url{https://github.com/sidpatelgit/polycomp} contains code and frozen manifests for the 120 problems, including exact image-hash and geometry validation. It also contains payload builders for all three presentations, as well as a table with the results from the evaluation of all three models.

\section{Limitations}

\begin{itemize}[leftmargin=*]
\item Perception might be a more challenging component of these problems than we expect. Alternate line weights, colors, or image resolutions could improve performance.
\item We did not test whether variations of the prompt improved performance.
\item We did not collect a formal human performance baseline. We did verify that each problem can be solved as rendered by rejecting all of the incorrect options, but some problems are still quite challenging.
\item We did not evaluate run-to-run repeatability in individual model responses or in aggregate performance.
\item We did not test whether test-time training or a harness would improve performance.
\item We used one model (GPT-5.5 with Extra High effort) to develop the problem generation procedure, to generate candidate problems, and to find hard problems (i.e., problems that GPT-5.5 did not solve or solved most slowly). This could bias evaluation if this single-model process tends to produce problems that are easier or harder for GPT-5.6 than for other models.
\end{itemize}

\section{Conclusion and extensions}

\textsc{PolyComp} identifies a gap in multimodal model spatial intelligence in a constrained setting: polycube geometry, rotation, and composition. It also presents a method for generating verifiable problems of this class.

Future directions include:

\begin{itemize}
\item further simplifying the perception task by identifying how small of a target solid (i.e., how few cubes in the polycube) still challenges models;
\item scaffolding the perception task by instructing the model to first create three-dimensional matrices representing the target solid and the component solids in each option;
\item using the polycube geometry to test understanding of silhouettes, cross sections, interference-free assembly, and other spatial reasoning tasks that may involve internal representations of three-dimensional objects;
\item isolating what types of disqualifying features for the wrong options the models are able to identify most easily;
\item moving from polycubes to similar smooth and rounded geometry;
\item using multiple models to generate and select problems within a common framework to reduce the risk of biased evaluation.
\end{itemize}

 \section*{Acknowledgments}

We thank Ranjay Krishna for providing guidance about single vs. multi-image presentations and the evaluation protocol.

\begingroup
\interlinepenalty=10000
\bibliographystyle{unsrtnat}
\bibliography{references}
\endgroup

\end{document}